# Physics of Language Models: Part 3.1, Knowledge Storage and Extraction

Zeyuan Allen-Zhu
zeyuanallenzhu@meta.com
Meta / FAIR Labs

Yuanzhi Li
Yuanzhi.Li@mbzuai.ac.ae
Mohamed bin Zayed University of AI

September 18, 2023

(version 3)*

**Abstract**

Large language models (LLMs) can store a vast amount of world knowledge, often extractable via question-answering (e.g., "What is Abraham Lincoln's birthday?"). However, do they answer such questions based on exposure to similar questions during training (i.e., cheating), or by genuinely learning to extract knowledge from sources like Wikipedia?

In this paper, we investigate this issue using a controlled biography dataset. We find a strong correlation between the model's ability to extract knowledge and various *diversity measures* of the training data. **Essentially**, for knowledge to be reliably extracted, it must be sufficiently augmented (e.g., through paraphrasing, sentence shuffling, translations) *during pretraining*. Without such augmentation, knowledge may be memorized but not extractable, leading to 0% accuracy, regardless of subsequent instruction fine-tuning.

To understand why this occurs, we employ (nearly) linear probing to demonstrate a strong connection between the observed correlation and *how the model internally encodes knowledge* — whether it is linearly encoded in the hidden embeddings of entity names or distributed across other token embeddings in the training text.

**This paper provides several key recommendations for LLM pretraining in the industry: (1) rewrite the pretraining data — using small, auxiliary models — to provide knowledge augmentation, and (2) incorporate more instruction-finetuning data into the pretraining stage before it becomes too late.**

---

*Project page: https://physics.allen-zhu.com/part-3-knowledge/part-3-1. An extended video of this paper is available at https://youtu.be/YSHzKmEianc. V1 was circulated internally at Meta on Sep 18, 2023, and appeared on arXiv on Sep 25, 2023. V2 is nearly identical to V1, with minor corrections to author names and writing. V3 includes additional Llama experiments and further writing improvements.

We would like to thank Lin Xiao, Chunting Zhou, Tianyi Peng, Xiaodong Liu, and Zhijie Zhou for many helpful conversations. We would like to extend special thanks to Nabib Ahmed, Giri Anantharaman, Lucca Bertoncini, Henry Estela, Liao Hu, Caleb Ho, Wil Johnson, Apostolos Kokolis, and Shubho Sengupta from Meta FAIR, as well as Ian Clark, Gourab De, Anmol Mann, and Max Pfeifer from W&B; without their invaluable support, the experiments in this paper would not have been possible.

# 1 Introduction

Knowledge is crucial for human cognition and communication, allowing us to comprehend and utilize information. For humans, this often involves memorization, the process of storing and retrieving information in the brain. For example, after reading a biography of Abraham Lincoln, we can memorize the information and later answer questions like "Where was Lincoln born?" or "What is Lincoln's birthday?" Memorization enables us to extract and manipulate knowledge from the sentences we read or hear, recognize the entities, relations, and facts expressed in the text, and apply logical and causal reasoning to infer new information or answer queries [4, 6, 12, 42].

In this paper, we explore **how transformer-based language models** memorize knowledge during training and extract it during inference. This is distinct from in-context learning or RAG [22], where the model is given a paragraph during inference and immediately answers questions about it. We focus on *factual knowledge* (e.g., knowledge graph) that a language model needs to memorize from the training corpus, encode in its weights, and extract later during inference.

We stress that *memorizing* all sentences in the training data **does not** ensure that the model can *extract or manipulate* the factual knowledge from the sentences during inference. Language models can reproduce the exact input during inference, but this doesn't necessarily mean they can use these sentences to answer factual questions related to them. Hence, we differentiate between "memorization of knowledge" in language models and traditional memorization in machine learning, which merely means the model can fit the exact training data, but doesn't imply the model can **extract the knowledge flexibly** from the data after training.

For example, if the training data includes Lincoln's biography, the model can memorize and reproduce the sentence "Abraham Lincoln was born in Hodgenville, K.Y." when given the prompt "Abraham Lincoln was born in", but it might not be able to answer the question "Which city was Abraham Lincoln born in?" Therefore, a key question is:

*How do language models memorize knowledge during training, and extract it later to answer questions or perform logical reasoning during inference?*

Previous works have demonstrated that language models can "memorize" a lot of knowledge by probing the model to answer questions related to different entities and attributes, see [28, 33, 35] and the citations therein. These studies use models pretrained on internet data, leaving it **unclear** whether the model answers questions like "Which city was Abraham Lincoln born in?" by *extracting knowledge* from Lincoln's biography (**our focus**) or if it encountered a similar (or same!) question during training and memorized the answer (traditional memorization / **data contamination**).

Given the challenges of conducting controlled experiments with internet data, we propose studying this question using well-controlled, synthetically generated data,[1] examining the models' mathematical properties that characterize their knowledge representation and extraction. We construct a synthetic dataset of $100k$ biographies, including their birthday, birth city, major of study, etc. We also use Llama [37] to rewrite them to make them close to real-life biography styles. We pretrain the language model on the biography dataset of all the $100k$ people. We ask:[2]

*After pretraining a language model on the biography dataset, can the model be finetuned to answer questions like "Where is the birth city of [name ]", and if so, how does the model achieve so?*

---

[1]One could suggest filtering the data to eliminate such questions and retraining the model. However, this doesn't rule out the presence of similar sentences "Which city did Abraham Lincoln grow up in?", more complex ones in French, or grammatically incorrect versions like "Where Abraham Lincoln birth in?" in the data.

[2]We leave the follow-up question to study *logical reasoning or manipulation* on knowledge to a separate paper [2].

After pretraining the model on the entire biography, we fine-tune it using question and answer (QA) pairs from a $p$ fraction of individuals. We then test its ability to *out-of-distribution* answer QAs about the remaining $1-p$ fraction. This approach ensures that the model (1) is exposed to sufficient data to comprehend the QAs and (2) does not encounter the same questions during training. The paper is structured as follows:[3]

- RESULT 1: MIXED TRAINING $\implies$ KNOWLEDGE EXTRACTION.

  Before diving into the pretrain-finetune process, we first demonstrate that pretraining a model on all biographies *plus* QAs for a $p$ fraction of individuals together enables it to (apply knowledge to) answer questions about the remaining $1-p$ fraction. We call this process *mixed training*. We observe in mixed training, the model *first uses QAs* to encode knowledge about the $p$ fraction, then correlates this encoded knowledge with the biography to infer generalization to the remaining $1-p$ fraction. This learning process deviates from typical human learning[4] and is less frequently used in practical LLM pretrain (and perhaps it should!).

- RESULT 2-3: INSTRUCT FINETUNE $\not\implies$ KNOWLEDGE EXTRACTION (UNLESS DATA AUGMENTED).

  Consider a model pretrained only on the biographies and then finetuned using QAs for a $p$ fraction of individuals. We discover that it struggles to answer questions about the remaining $1-p$ fraction, *irrespective of model size, pre-train time, or finetune parameters* (Result 2). However, accuracy significantly improves with *knowledge augmentations* like varying writing styles or sentence shuffling (Result 3). This gives **a strong link** between knowledge augmentation in the pretrain data and the model's knowledge extraction ability after finetuning.

- RESULT 4-5: INTRODUCE PROBING TECHNIQUES TO EXPLAIN <u>WHY THIS HAPPENS</u>.

  As another main contribution, we introduce (nearly) linear probing techniques to show that knowledge augmentation pushes the model to encode a person's knowledge almost linearly in the model's hidden embedding of the person's name tokens. Without augmentation, the model encodes the person's knowledge across all biography words/tokens, making knowledge extraction nearly impossible no matter how one finetunes it. In sum:

  no knowledge augmentation in pretrain data $\iff$ attribute is **not** entirely stored on person's names
  
  when the model memorizes the pretrain data
  
  $\iff$ knowledge cannot be extracted via instruction finetune
  
  knowledge augmented in pretrain data $\iff$ attribute is **nearly** entirely stored on person's names
  
  $\iff$ knowledge can be extracted via instruction finetune

- RESULT 6: KNOWLEDGE AUGMENTATION ON THE "CELEBRITY" HELPS "MINORITY".

  Even if knowledge augmentation is applied to a subset of individuals, what we call celebrities, test accuracy for others (without augmentation) also increases significantly. We discover that the mere inclusion of celebrity data (e.g., people with plentiful online biographical data of diverse writing styles) in pre-training enhances the model's knowledge extraction for minorities.

- RESULT 7: BI-DIRECTIONAL MODELS FAIL TO EXTRACT KNOWLEDGE.

  We show that *encoder-only models akin to BERT*, whether mixed-trained or pre-trained and then fine-tuned, cannot extract a person's knowledge after finetuning, regardless of the knowledge augmentation, unless the knowledge is a single word or multiple but independent words (like birth month, day, and year).

---

[3]Our result numbers correspond to our online video of the paper, available at `https://youtu.be/YSHzKmEianc`.

[4]For humans, arguably, we first learn from textbooks and then answer exam questions.

**Practical Implications.** Our controlled study offers key recommendations for LLM training at an industrial scale:

- We emphasize the **importance of pre-training data rewriting (augmentation)**, particularly for rare but critical data. Addressing this during fine-tuning is often too late. Without rewriting, a model may accurately recite knowledge data word by word, but the way it embeds this knowledge into its weights may impede retrieval when prompted differently, resulting in a *total waste of model capacity*.

  Tools such as Llama-7B or even *smaller* auxiliary models are adequate for this rewriting task. These "rewrite models" do not need to possess the knowledge themselves. As demonstrated, simple sentence-level shuffling or English-to-French translations can already enhance performance. Generally, we suggest including prompts that encourage sentence shuffling when using such rewrite models.

  Data rewriting is a form of data augmentation, but also distinct from traditional methods (e.g., dropout, masking, cropping, jittering, flipping) and their associated distillation techniques (like contrastive learning). While traditional augmentations promote the learning of generalizable features over pure memorization, data rewriting — what we call knowledge augmentation — helps language models to memorize knowledge in a more accessible format for downstream tasks. Without such augmentation, the accuracy even for the simplest knowledge extraction task, could be near zero.

- We also demonstrate the advantages of **including more instruction-finetuned data during pre-training**. Our mixed training experiments show that postponing all QA-like data to the fine-tuning phase is suboptimal. Introducing QA-like data earlier in pre-training enables the model to *encode knowledge more effectively*.

## 1.1 Related Work

LINEAR PROBING OF KNOWLEDGE. Linear probing is a recognized method to examine how a model encodes knowledge [5, 11, 13, 15, 23, 26, 35]. Contrary to previous studies that suggest models trained on internet data can linearly encode knowledge in the hidden embeddings of entity names, we find that such encoding is only possible with knowledge augmentations like permutation/rewriting of entity-attribute knowledge during pretraining. Without these augmentations, the language model can still memorize the training data, but it is not linearly encoded in the entity's hidden embeddings, making knowledge extraction via QAs quite hard, if not impossible, even with instruction finetuning. This implies that diverse internet data on the same entity is vital for pre-training the language model for knowledge extraction during inference. The usefulness of augmentations of pretraining data for language models was also empirically observed in literature [7, 9, 14, 21], but they did not explore **where** the knowledge is nearly-linearly encoded in a sentence and its correlation with knowledge augmentation, a process we refer to as P-probing in Section 5.1.

PROBING LANGUAGE MODELS' KNOWLEDGE VIA QAS. Question answering (QA) is a common method to probe the knowledge encoded in language models pretrained on internet data [17, 27–30, 32, 33, 35]. However, it's unclear whether these models answer questions by extracting knowledge from the training source or by recognizing exact/similar questions from training. We use controlled experiment for out-of-distribution testing on individuals whose QAs were not part of training. This approach also allows us to study the correlation between knowledge extraction and the diversity of pretrain data.

ENCODER VERSUS DECODER FOR QAS. While BERT-based models [20] are also used for knowledge extraction through QAs [10, 36], our work indicates that they are less effective at extracting knowledge compared to GPT models.

## 2   Result 0: Our Dataset Families

In this paper, we introduce synthetic human biography datasets and near-real datasets generated by LLaMa [37, 40]. Detailed descriptions are in the appendix, with a brief overview here.

**BIO dataset bioS.** The synthetic dataset, bioS, generates profiles for $N = 100,000$ individuals.[5] Each individual's details are randomly and *independently* selected from a uniform distribution. The birth dates offer $200 \times 12 \times 28$ possibilities, while other categories offer $100 \sim 1,000$ choices. We also add a "company city" attribute which *depends* on the employer's headquarters location. We ensure uniqueness in each individual's full name.

We generate a six-sentence biographical text entry for each individual, highlighting six distinct aspects. For diversity, each sentence is randomly chosen from approximately 50 distinct templates. In the basic configuration, we generate a single biographical entry for each person, maintaining a consistent order for the six sentences. We use "bioS single" to denote this basic configuration. See an example entry below:

> Anya Briar Forger was born on October 2, 1996. She spent her early years in Princeton, NJ. She received mentorship and guidance from faculty members at Massachusetts Institute of Technology. She completed her education with a focus on Communications. She had a professional role at Meta Platforms. She was employed in Menlo Park, CA.

(2.1)

We also explore 3 types of knowledge augmentations: (1) multi$M$, generating $M$ biography entries for an individual using varied templates, (2) fullname, substituting he/she/they with the person's full name; and (3) permute, shuffling the six sentences randomly. Examples are given in Section 4.2.

**BIO dataset bioR.** We examine a "close-to-real" dataset produced by Llama [37, 40]. For the set of $N = 100,000$ individuals, we provide an instructive prompt to Llama to generate a biographical entry. Here's an example:

> Anya Briar Forger is a renowned social media strategist and community manager. She is currently working as a Marketing Manager at Meta Platforms. She completed her graduation from MIT with a degree in Communications. She was born on 2nd October 1996 in Princeton, NJ and was brought up in the same city. She later moved to Menlo Park in California to be a part of Facebook's team. She is an avid reader and loves traveling.

We diversified our instructive prompts by drawing from a pool of templates and employed rejection sampling to guarantee the inclusion of all six attributes. In the basic configuration, we produce a single biographical entry for each person (denoted as "bioR single"). For comparison, we also consider multi$M$ augmentation which generates $M$ entries per person and the fullname augmentation. Additional examples can be found in Appendix A.

**QA dataset.** This paper explores the effectiveness of a trained language model in retaining knowledge from BIO data. As discussed in the introduction, memorization *is more than just predicting the next token* when given exact sentences from BIO. It includes the model's ability to truly **extract knowledge from the BIO**. We assess this knowledge extraction using a question and answer (QA) framework. For each individual, we pose six questions targeting their six unique attributes:

---

[5]We have a follow-up to push this to $N = 20,000,000$ and similar results hold [3].

1. What is the birth date of Anya Briar Forger?
   Answer: October 2, 1996.
2. What is the birth city of Anya Briar Forger?
   Answer: Princeton, NJ.
3. Which university did Anya Briar Forger study?
   Answer: Massachusetts Institute of Technology.
4. What major did Anya Briar Forger study?
   Answer: Communications.
5. Which company did Anya Briar Forger work for?
   Answer: Meta Platforms.
6. Where did Anya Briar Forger work?
   Answer: Menlo Park, CA.

For each question, we use it as a prompt for the model to generate a response. QA accuracy is measured by the proportion of answers that exactly match the correct response.[6]

## 2.1 Training Details

**Model architectures.** We adopt the GPT2/Llama architectures [31, 37], where for GPT2 we replace its absolute positional embedding with *rotary positional embedding* [8, 34], but still referring it as GPT2 for short.[7] Recall the GPT2-small architecture comprises 12 layers with 12 heads and 768 dimensions [31]. We use a 12-layer, 12-head, 768-dim GPT2 (124M) or Llama architecture for pretraining on the bioS data, but a larger 12-layer, 20-head, 1280-dim GPT2 (302M) or Llama for the bioR data to accommodate its increased complexity. Only in Figure 2 when presenting a negative result, we tried a 12-layer 32-head 2048-dim GPT2 (682M). The default GPT2/Llama tokenizers are used, which convert simple words into single tokens, but names and most other attributes into tokens of varying lengths. When it comes to Section 7, we also use a BERT architecture [20].

**Training.** We investigate two types of autoregressive training, detailed in Appendix B.

<u>Pretrain + instruction finetune</u>. Here, we pre-train the language model ***from scratch*** on the BIO data, randomly sampling and concatenating them into 512-token sentences, separated by a standard `<EOS>` token. The model is then fine-tuned using half of the QA data and evaluated on the remaining half, mirroring the typical instruction finetune process.

<u>Mixed training</u>. In mixed training, we train the model ***from scratch*** on all BIO data and half of the QA data. BIO and QA entries are randomly sampled without requiring them to be from the same individual. We use a parameter $QA_r$ to control the QA data amount, primarily setting $QA_r = 0.8$ (a 2 : 8 BIO to QA entry ratio). The model's generation accuracy is evaluated using the remaining QA data.[8]

**LoRA + full finetune.** In full finetuning a pretrained model is tuned for a downstream task such as QAs. LoRA finetuning [18] improves upon this by freezing all pretrained model parameters and adding low-rank updates to a subset of the weight matrices for fine-tuning. We apply a low-rank update to the query/value matrices of the transformer model (suggested by [18]) and the embedding layer to account for input data distribution shifts. *Full finetuning is also included* when presenting negative results.

## 3 Result 1: Mixed Training Enables Knowledge Extraction

Mixed training involves using BIO data for *all* individuals together with QAs for half of them. The group of individuals whose QAs are included in the training set is referred to as *in-distribution* or $\mathcal{P}_{\mathsf{train}}$. The model's generative accuracy is then tested on the QAs from the remaining individuals ($\mathcal{P}_{\mathsf{test}}$) to assess its out-of-distribution (OOD) generalization capability.

---

[6] We disregard partial matches or synonyms, emphasizing the model's precision in knowledge extraction.

[7] A controlled experiment to highlight the importance of rotary embedding is in [1]. Since this paper appeared, Jiang et al. [19] confirms our results also apply to the pretrained Llama-7B model; our own follow-up also tried the Mistral architecture [3].

[8] See Appendix C for a comparison of how $QA_r$ affects performance. We used beam =4 without sampling throughout this paper; results are similar if disabling beam.

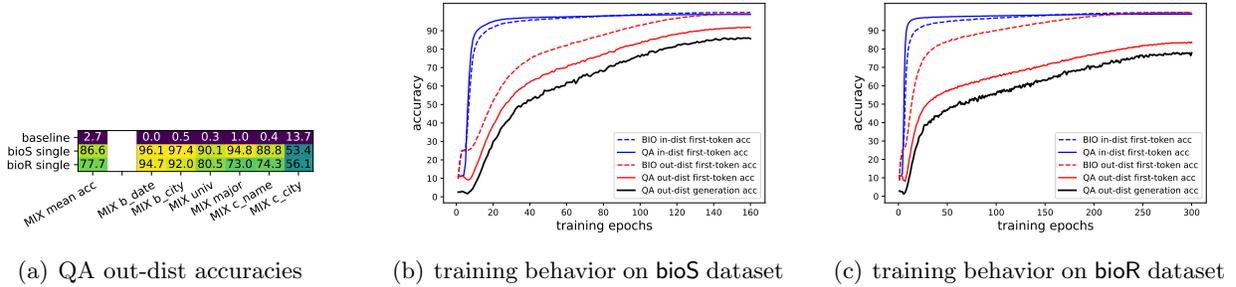

(a) QA out-dist accuracies  (b) training behavior on bioS dataset  (c) training behavior on bioR dataset

Figure 1: Accuracies and loss curves for mixed training (GPT2). b_date,b_city,c_name,c_city stand for birth date, birth city, company name, company city, and mean acc stands for the mean accuracy of the six attributes. Baseline is majority-guessing (c_city has large accuracy because many companies are based in NYC).

> **Result 1** (Figure 1). *A model mixed-trained on both knowledge and its extraction QA tasks can effectively learn to extract knowledge.*
>
> - *As shown in Figure 1(a), the OOD generalization accuracies are* 86.6% *when mixed-trained on bioS single and* 77.7% *for bioR single.*
> - *However, the model achieves this through somewhat abnormal behavior akin to "studying to pass the test," discussed further in Section 3.1.*

(We emphasize that the accuracy is OOD: extracting an individual's attributes even when no QA about that person — and only the BIO of that person — was seen in the training data.)

## 3.1 Model's Abnormal Learning Behavior

We examine the model's mixed training for knowledge storage and extraction by monitoring its accuracies on the BIO/QA data and for $\mathcal{P}_{\mathsf{train}}/\mathcal{P}_{\mathsf{test}}$ separately. Specifically,[9]

- BIO first-token accuracy: we track the model's next-token-prediction accuracy on the first token of each of the six attributes (birthdate, birthcity, etc.) in the BIO data, separately for $\mathcal{P}_{\mathsf{train}}/\mathcal{P}_{\mathsf{test}}$. This measures the model's BIO data memorization performance. (Despite all individuals' BIO data appearing in training, we still separately track them for $\mathcal{P}_{\mathsf{train}}/\mathcal{P}_{\mathsf{test}}$.)
- QA first-token accuracy: we track the model's next-token-prediction accuracy on the first answer token in the QA data, separately for $\mathcal{P}_{\mathsf{train}}/\mathcal{P}_{\mathsf{test}}$. This loosely estimates the model's QA generation performance.
- QA generation accuracy: we track the model's whole-attribute generation accuracy on $\mathcal{P}_{\mathsf{test}}$.

From Figure 1(b) and 1(c), we find that the model employs an unconventional learning strategy.

- Initially, the model uses the QA data from the training set to encode knowledge for people in $\mathcal{P}_{\mathsf{train}}$, as indicated by the rapid increase in QA in-dist accuracy. This also aids in memorizing in-dist BIO data, as shown by the subsequent rise of the BIO in-dist accuracy.
- The model then gradually aligns the encoded knowledge with the BIO data to learn to extract knowledge and generalize it to $\mathcal{P}_{\mathsf{test}}$. Notably, it takes a while before the BIO out-dist accuracy

---
[9]Interested readers may consider "whole-attribute" accuracies instead of "first-token" accuracies. They are similar, so we omit them here.

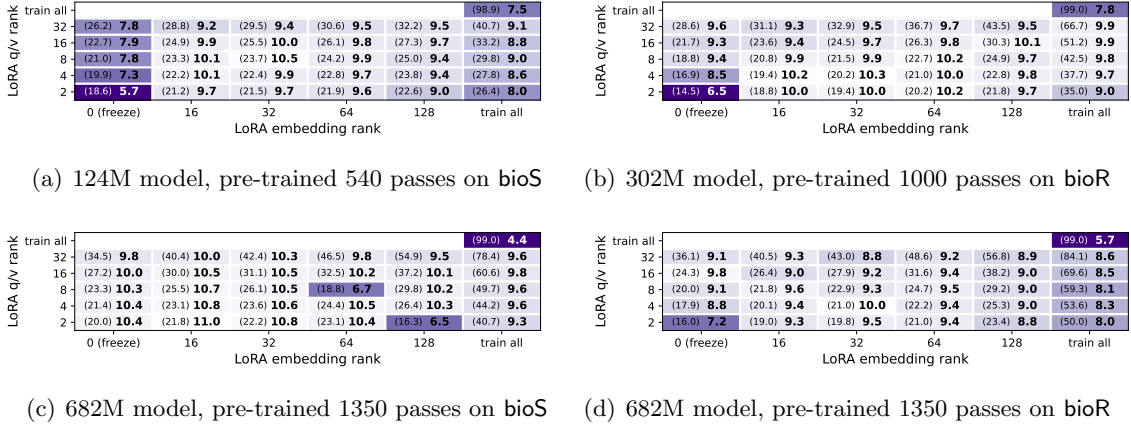

(a) 124M model, pre-trained 540 passes on bioS

(b) 302M model, pre-trained 1000 passes on bioR

(c) 682M model, pre-trained 1350 passes on bioS

(d) 682M model, pre-trained 1350 passes on bioR

Figure 2: BIO pretrain + QA finetune (train acc) / **test acc** using GPT2. Bold number indicates QA generation accuracy on $\mathcal{P}_{\text{test}}$, and the smaller number in parentheses represents QA (first-token) accuracy on $\mathcal{P}_{\text{train}}$. For **LoRA fine-tune** we consider a rank $r = 2, 4, 8, 16, 32$ update on the query/value (q/v) matrices and a rank $r' = 0, 16, 32, 64, 128$ update on the word embedding matrix. **Full finetune** is included in the upper-right corners (train all / train all). More details are in Appendix D.

catches up, followed by an increase in the QA out-dist accuracy.

This is akin to the "study to pass the test" approach in schools, where students prepare using past exam questions and textbooks for answers. While this may yield high scores, it doesn't reflect the natural progression of human knowledge acquisition. **To address this**, we explore a more challenging scenario in the next section where the model is pretrained on the BIO data without exposure to the questions.

*Remark* 3.1. In mixed training, we selected $\mathsf{QA}_r = 0.8$, maintaining a $8:2$ QA to BIO ratio as outlined in Section 2. We found that a higher QA ratio during training improved out-of-distribution QA accuracy (Figure 10 in Appendix C), further supporting our observation of the model's abnormal behavior: it first learns knowledge from QA and then associates it with BIO. For comparison, Llama was trained using only 2% of tokens from StackExchange [37].

## 4 Result 2-3: BIO Pretrain + QA Instruction Finetune

We explore a scenario where the model is pre-trained exclusively on the BIO data of all individuals, followed by fine-tuning using QAs from half of these individuals, denoted as $\mathcal{P}_{\text{train}}$. The model's OOD generalization is then assessed on questions related to the other half, denoted as $\mathcal{P}_{\text{test}}$, whose BIO/QA data were not involved in the fine-tuning. This setup simulates the process of applying learned knowledge from textbooks to solve exam questions.

### 4.1 Result 2: Model Fails to Extract Knowledge After BIO Pretrain

We first pretrain on bioS or bioR single, each containing a single biography per person. The QA finetune generalization accuracies (on $\mathcal{P}_{\text{test}}$) are shown in Figure 2, using both full and LoRA finetuning [18]. The model's QA finetune training accuracy on $\mathcal{P}_{\text{train}}$ is also presented for comparison.

Despite a 99+% first-token accuracy during pretraining, the model exhibits zero-zero QA accuracy on $\mathcal{P}_{\text{test}}$ for all finetuning parameters. This indicates that while the model can memorize BIO data token-by-token, it struggles to extract the underlying knowledge. Full-finetuning yields

| | QA mean acc | QA b_date | QA b_city | QA univ | QA major | QA c_name | QA c_city | MIX mean acc | MIX b_date | MIX b_city | MIX univ | MIX major | MIX c_name | MIX c_city |
|---|---|---|---|---|---|---|---|---|---|---|---|---|---|---|
| baseline | 2.7 | 0.0 | 0.5 | 0.3 | 1.0 | 0.4 | 13.7 | 2.7 | 0.0 | 0.5 | 0.3 | 1.0 | 0.4 | 13.7 |
| bioS single | 9.7 | 33.5 | 6.3 | 2.3 | 4.0 | 1.1 | 13.8 | 86.6 | 96.1 | 97.4 | 90.1 | 94.8 | 88.8 | 53.4 |
| bioS single + fullname | 48.9 | 56.2 | 58.8 | 63.0 | 55.7 | 50.5 | 14.1 | 85.9 | 95.8 | 97.7 | 88.7 | 94.4 | 86.0 | 55.9 |
| bioS single + permute1 | 4.4 | 0.5 | 3.3 | 2.4 | 5.0 | 3.5 | 13.7 | 82.5 | 92.2 | 94.5 | 86.4 | 87.4 | 70.2 | 67.2 |
| bioS single + permute2 | 53.2 | 57.3 | 48.3 | 53.1 | 55.0 | 51.8 | 58.3 | 91.6 | 95.7 | 97.8 | 89.6 | 92.1 | 88.6 | 89.2 |
| bioS single + permute5 | 70.0 | 56.4 | 57.7 | 58.3 | 64.9 | 90.5 | 97.7 | 93.7 | 97.0 | 97.4 | 89.7 | 91.6 | 92.2 | 96.5 |
| bioS single + permute1 + fullname | 31.7 | 26.6 | 29.3 | 36.9 | 31.1 | 31.4 | 37.9 | 89.8 | 94.9 | 97.4 | 89.7 | 90.7 | 84.0 | 84.7 |
| bioS single + permute2 + fullname | 73.1 | 69.0 | 60.6 | 64.2 | 64.0 | 87.9 | 95.0 | 92.6 | 95.6 | 98.1 | 89.2 | 91.5 | 90.6 | 93.4 |
| bioS single + permute5 + fullname | 80.2 | 83.7 | 67.8 | 72.6 | 69.1 | 93.0 | 98.6 | 93.4 | 95.1 | 97.9 | 88.9 | 92.7 | 90.7 | 97.4 |
| bioS multi2 | 41.1 | 100 | 71.7 | 33.1 | 26.1 | 5.2 | 14.0 | 89.2 | 99.4 | 98.3 | 89.6 | 96.6 | 92.2 | 61.3 |
| bioS multi2 + fullname | 84.0 | 100 | 97.7 | 89.5 | 97.6 | 91.3 | 35.3 | 87.9 | 99.8 | 98.8 | 88.6 | 96.6 | 87.6 | 58.0 |
| bioS multi2 + permute | 91.2 | 99.3 | 98.7 | 89.8 | 96.7 | 83.3 | 83.5 | 91.6 | 98.1 | 97.6 | 88.1 | 96.2 | 87.2 | 85.4 |
| bioS multi2 + permute + fullname | 96.1 | 100 | 98.8 | 91.3 | 98.1 | 93.7 | 97.8 | 94.4 | 99.3 | 98.6 | 89.7 | 96.6 | 92.2 | 92.6 |
| bioS multi5 | 41.0 | 100 | 50.8 | 30.9 | 43.5 | 10.2 | 13.8 | 91.8 | 99.9 | 99.0 | 91.1 | 97.2 | 93.7 | 71.7 |
| bioS multi5 + fullname | 82.4 | 100 | 98.6 | 88.4 | 96.1 | 91.9 | 26.8 | 92.0 | 99.9 | 98.7 | 91.0 | 97.4 | 93.2 | 74.6 |
| bioS multi5 + permute | 96.6 | 100 | 99.0 | 91.3 | 97.7 | 95.1 | 98.7 | 95.5 | 99.8 | 98.1 | 90.0 | 97.4 | 93.7 | 96.8 |
| bioS multi5 + permute + fullname | 96.2 | 100 | 98.7 | 90.6 | 97.9 | 93.7 | 99.0 | 95.7 | 99.8 | 98.7 | 89.5 | 97.4 | 93.2 | 97.9 |

Figure 3: Comparison of BIO Pretraining + QA Finetuning (**left**) versus their Mixed Training counterparts (**right**) under various knowledge augmentations on the data (the **rows**). Displayed values indicate QA generation accuracies for six attributes in $\mathcal{P}_{\text{test}}$. This figure is for the GPT2 model on the bioS data; refer to Figure 12 for similar results on the bioR data and/or using the Llama architecture, and Appendix D for more details.

---

**Observation.** Knowledge augmentation in pretraining data improves model generalization to out-of-distribution QAs after finetuning. Accuracy increases with more augmentations introduced; while mixed training is minimally impacted by knowledge augmentation.

---

near-perfect *in-distribution* QA accuracy on $\mathcal{P}_{\text{train}}$, showing it can memorize QAs for individuals in the fine-tuning set. However, it fails to generalize to QAs about individuals in $\mathcal{P}_{\text{test}}$. In sum:

> **Result 2** (Figure 2). *A model pretrained to word-by-word memorize knowledge may* **never be fine-tuned** *to extract knowledge. As shown in Figure 2:*
>
> *perfect BIO token memorization + perfect QA answers for half the people*
> $\not\Longrightarrow$ *correct QA answers for the other half.* (*knowledge extraction does not come for free*)

This holds true even when the model size is $\sim 7000\text{x}$ larger than $N = 100k$, with each individual exposed 1350 times during pretraining, and numerous finetuning parameters have been explored.[10] Despite memorizing all knowledge from the BIO data during pretraining, the model encodes it in a disorganized manner within the transformer, preventing knowledge extraction during fine-tuning.[11]

Figure 2 seems to contradict the success of large models like GPT3.5/4, trained on diverse internet data such as Common Crawl and known for effective knowledge extraction upon fine-tuning. Analyzing the test accuracy breakdown for the six attributes on the bioS data (Figure 3, the "bioS single" row), we find that QA fine-tuning achieves a 33% generalization accuracy on the "birthdate" attribute but performs poorly on others. This is because our bioS single data consistently places birthdate as the first attribute after a person's name, unlike internet data which presents information variably, often repeating it with diverse wordings and orderings. The next subsection on knowledge augmentation supports this hypothesis.

## 4.2 Result 3: Knowledge Augmentation

We explore how knowledge augmentation enhances a model's capacity to store and efficiently extract knowledge from training data. We focus on three augmentations: adding multiplicity, introducing permutations, and repeating full names, typically found in internet data. The original datasets without augmentation are referred to as bioS single and bioR single.

---

[10]In our follow-up work [3], we increase the model size to 1B and $N$ to 20M, confirming similar results.

[11]This is not a direct result of catastrophic forgetting, a common issue during heavy fine-tuning where the model forgets the pretraining data. Even with LoRA fine-tuning, which introduces minimal low-rank updates to model weights while preserving the pretrained model, test accuracy only slightly improves.

- MULTIPLICITY. We denote the method of creating $M$ distinct biography entries for each individual, using varied language but retaining the same information, as multi$M$.[12] An example of adding multiplicity to the biography in (2.1) is:

  *Anya Briar Forger* came into this world on *October 2, 1996*. She originated from *Princeton, NJ*. She pursued advanced coursework at *Massachusetts Institute of Technology*. She dedicated her studies to *Communications*. She developed her career at *Meta Platforms*. She gained work experience in *Menlo Park, CA*.

  *Remark.* As a special case, we also experimented with translation (e.g., English to French) to increase sentence diversity, which proved beneficial for the model's knowledge extraction, but we have not included these details in this paper for clarity.

- PERMUTATION. We denote adding random permutations to the biography sentences as permute.[13] For instance, the example above can be permuted as follows:

  *Anya Briar Forger* originated from *Princeton, NJ*. She dedicated her studies to *Communications*. She gained work experience in *Menlo Park, CA*. She developed her career at *Meta Platforms*. She came into this world on *October 2, 1996*. She pursued advanced coursework at *Massachusetts Institute of Technology*.

- FULLNAME. We denote the augmentation where all pronouns or partial names in bioS/bioR are replaced with the person's full name as fullname. [14] An example of this augmentation is:

  *Anya Briar Forger* originated from *Princeton, NJ*. Anya Briar Forger dedicated her studies to *Communications*. Anya Briar Forger gained work experience in *Menlo Park, CA*. Anya Briar Forger developed her career at *Meta Platforms*. Anya Briar Forger came into this world on *October 2, 1996*. Anya Briar Forger pursued advanced coursework at *Massachusetts Institute of Technology*.

**Results.** In Figure 3, we present our results for the GPT2 model on the bioS dataset. We implemented each knowledge augmentation individually and in combinations, then compared the model's QA finetune accuracy on $\mathcal{P}_{\text{test}}$. The model architecture and training parameters remained the same, but the pre-training datasets varied based on the applied augmentations. Further experiment details are in Appendix D; and additional results for the bioR dataset and/or Llama model are in Figure 12. We find that:

> **Result 3** (Figure 3). *Adding multiplicity, permutations, or repeating full names, all help the model to **better store knowledge** during pretraining, making knowledge extraction easier later. Notably:*[15]
> - *Pretraining on a dataset where each person has 5 diverse biography entries (i.e., different wording, sentence shuffling) boosts the QA fine-tune accuracy (on $\mathcal{P}_{\text{test}}$) from 9.7% to 96.6%.*
> - *More augmentation $\Rightarrow$ better: gain increases as multiplicity or permutation number increases.*

One may infer from Result 3 that exposing the model to varied expressions of the same knowledge encourages it to focus on the underlying structure of the knowledge, rather than its word-by-word presentation. We shall verify this hypothesis in Section 5 by introducing probing techniques.

---

[12]For bioS data, each of the six sentences is selected from around 50 templates, with a new template resampled for each sentence in the $M$ entries. For bioR data, we recreate the biography using Llama for each of the $M$ entries.

[13]For bioS single, we denote random permutation of the same six sentences $P$ times as permute$P$. For bioS multi$M$, we denote random permutation of each of the $M$ biography entries as permute. The bioR data, generated by Llama, already has some randomness in sentence ordering, so no extra permutations are added.

[14]In the synthetic bioS dataset, a person's full name is presented only once, at the start of the initial sentence, with subsequent sentences using solely pronouns. For the LLaMa-generated bioR data, typically, the person's full name appears once at the start; later sentences use either pronouns or parts of the name, such as the first or last name.

[15]We have also tried to translate from English to French, which boosts accuracy to about 40% but we did not include the result for clarity. An exception is when permutation is directly added to the single data without multiplicity (see "bioS single + permute1"), this hurts the QA performance as it makes knowledge extraction harder.

# 5 Results 4-5: Knowledge Probes on the BIO Pretrained Model

We investigate how a language model, *pretrained on BIO data*, encodes knowledge in its hidden states. We propose two probing methods: position-based (P-probing) and query-based (Q-probing). Both methods employ simple, nearly-linear probes to extract personal attributes from the model's hidden states.

## 5.1 Result 4: Position-Based Probing

In P-probing, we feed biography entries into a pretrained model, and finetune an additional linear classifier on the model's final hidden layer to predict the six target attributes (e.g., university, major, etc.). We wish to understand how and where these attributes are encoded after pretraining.

To accommodate varied data lengths, we identify six *special token positions* immediately preceding the first occurrences of the six attributes in each biography entry (see Figure 4). This results in $6 \times 6$ classification tasks. For each prediction task, we freeze the entire pretrained network but add a trainable rank-2 update on the embedding layer to accommodate the task change. We use the transformer's last hidden layer at these positions to (linearly) predict the six target attributes.[16]

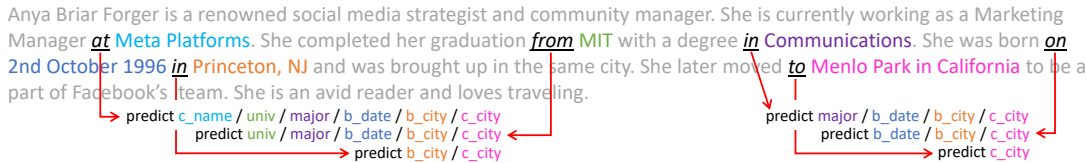

Figure 4: Illustration of the P-probing. Underscore prepositions are the *special token positions* where we prob. The task is to predict all attributes following these positions. Given the attribute ordering, there can be up to $6 \times 6 = 36$ tasks across all data.

We are particularly interested in how early the attributes are encoded in a biography. For instance, if the linear classifier to predict "company name" shows high accuracy right after the person's full name, it implies that the model is directly learning "Anya's employer is Meta Platforms". If high accuracy is only achieved at the biography's end, the model might be using a **flawed logic**, such as "the birthday is October 2, 1996, the university is MIT, hence the employer is Meta."

**P-Probing Main Results.** Our results are in Figure 5 and summarized as follows.

- In the bioS single setup, P-probing accuracy remains low (e.g., 2% for company name) until the token immediately preceding the target attribute (where accuracy boosts to 100%). This suggests that the model memorizes all the BIO data during pretraining, but encodes knowledge using the "flawed logic" above. This **prevents knowledge extraction** during QA finetuning, especially when only the person's name is provided.

- In the heavily augmented setup like bioS multi5+permute, the P-probing accuracy for all six attributes rises to nearly 100% from the first special position, which is before *all* of the attributes. This indicates that the model not only memorizes the BIO data but also identifies the person's complete six attributes solely upon seeing the person's name, **facilitating knowledge extraction** during the QA finetuning process.

---

[16]For GPT2-small with 768 hidden dimensions and vocab size $V$, this rank-2 update has $2V + 2 \times 768$ trainable parameters. The linear classifier layer is of dimension $768 \times M$ for each target attribute with $M$ possibilities. More details can be found in Appendix E.

| model | 0-bmonth | 1-bmonth | 2-bmonth | 3-bmonth | 4-bmonth | 5-bmonth | 0-bcity | 1-bcity | 2-bcity | 3-bcity | 4-bcity | 5-bcity | 0-univ | 1-univ | 2-univ | 3-univ | 4-univ | 5-univ | 0-major | 1-major | 2-major | 3-major | 4-major | 5-major | 0-cname | 1-cname | 2-cname | 3-cname | 4-cname | 5-cname | 0-ccity | 1-ccity | 2-ccity | 3-ccity | 4-ccity | 5-ccity |
|---|---|---|---|---|---|---|---|---|---|---|---|---|---|---|---|---|---|---|---|---|---|---|---|---|---|---|---|---|---|---|---|---|---|---|---|---|
| baseline | 8.3 | 8.3 | 8.3 | 8.3 | 8.3 | 8.3 | 2.5 | 2.5 | 2.5 | 2.5 | 2.5 | 2.5 | 37.0 | 37.0 | 37.0 | 37.0 | 37.0 | 37.0 | 4.0 | 4.0 | 4.0 | 4.0 | 4.0 | 4.0 | 1.5 | 1.5 | 1.5 | 1.5 | 1.5 | 1.5 | 14.8 | 14.8 | 14.8 | 14.8 | 14.8 | 14.8 |
| bioS single | 100 | | | | | | 5.9 | 100 | | | | | 38.0 | 37.1 | 99.2 | | | | 4.7 | 4.6 | 5.4 | 99.9 | | | 1.5 | 1.2 | 1.3 | 2.4 | 99.5 | | 15.4 | 15.4 | 14.9 | 13.0 | 69.1 | 100 |
| bioS single + fullname | 100 | | | | | | 52.1 | 100 | | | | | 67.3 | 74.2 | 99.9 | | | | 51.9 | 56.3 | 58.4 | 99.8 | | | 47.2 | 53.1 | 52.5 | 55.5 | 99.3 | | 28.2 | 31.3 | 30.9 | 32.9 | 75.8 | 99.9 |
| bioS single + permute1 | 26.1 | 28.9 | 32.8 | 40.5 | 56.8 | 100 | 19.2 | 22.9 | 28.5 | 36.3 | 53.3 | 100 | 47.5 | 50.1 | 53.0 | 57.6 | 69.2 | 98.8 | 20.2 | 24.2 | 28.8 | 37.6 | 55.5 | 100 | 21.4 | 33.0 | 45.5 | 57.9 | 72.6 | 98.3 | 27.1 | 45.3 | 62.2 | 78.1 | 92.2 | 100 |
| bioS single + permute2 | 87.7 | 88.8 | 90.1 | 91.6 | 94.0 | 100 | 52.1 | 55.5 | 60.0 | 64.7 | 73.7 | 100 | 65.6 | 67.8 | 70.6 | 74.6 | 80.7 | 99.9 | 61.1 | 64.4 | 68.5 | 72.4 | 80.6 | 100 | 61.6 | 70.1 | 77.3 | 84.7 | 91.9 | 99.9 | 66.9 | 76.1 | 83.7 | 91.1 | 97.1 | 100 |
| bioS single + permute5 | 96.1 | 96.3 | 96.7 | 97.1 | 97.9 | 100 | 58.0 | 60.8 | 63.7 | 67.8 | 76.8 | 100 | 71.5 | 73.1 | 74.8 | 77.9 | 84.1 | 99.9 | 72.5 | 74.5 | 76.3 | 79.2 | 84.7 | 100 | 97.0 | 97.1 | 97.3 | 97.9 | 98.6 | 100 | 99.3 | 99.5 | 99.6 | 99.8 | 99.9 | 100 |
| bioS single + permute1 + fullname | 58.8 | 64.3 | 69.6 | 74.4 | 82.9 | 100 | 37.4 | 41.6 | 47.9 | 56.1 | 69.7 | 99.9 | 54.9 | 59.1 | 64.0 | 70.1 | 79.0 | 98.9 | 42.0 | 47.2 | 52.7 | 60.1 | 71.8 | 100 | 43.2 | 54.2 | 65.3 | 76.8 | 88.3 | 99.8 | 49.5 | 61.8 | 74.6 | 85.1 | 95.6 | 100 |
| bioS single + permute2 + fullname | 81.5 | 85.0 | 86.7 | 88.2 | 92.1 | 100 | 57.7 | 63.2 | 65.9 | 71.1 | 78.2 | 100 | 69.7 | 72.4 | 75.5 | 78.0 | 83.6 | 99.7 | 65.3 | 69.6 | 72.8 | 76.6 | 82.2 | 100 | 91.9 | 93.9 | 94.8 | 96.0 | 97.4 | 100 | 96.3 | 97.4 | 98.2 | 98.8 | 99.6 | 100 |
| bioS single + permute5 + fullname | 88.8 | 90.4 | 91.5 | 92.3 | 94.6 | 100 | 63.5 | 67.3 | 69.9 | 73.6 | 80.4 | 100 | 76.8 | 80.0 | 81.8 | 83.8 | 88.1 | 99.9 | 70.4 | 72.9 | 75.1 | 78.2 | 83.9 | 100 | 98.0 | 98.0 | 98.3 | 98.7 | 99.0 | 100 | 99.9 | 100 | 100 | 100 | 100 | 100 |
| bioS multi2 | 100 | | | | | | 70.7 | 100 | | | | | 47.8 | 74.8 | 99.9 | | | | 18.9 | 30.1 | 60.1 | 99.6 | | | 3.0 | 3.8 | 8.4 | 34.6 | 99.3 | | 15.0 | 14.6 | 13.9 | 21.8 | 66.9 | 100 |
| bioS multi2 + fullname | 100 | | | | | | 100 | 100 | | | | | 99.6 | 100 | 100 | | | | 99.7 | 99.9 | 100 | 100 | | | 99.6 | 99.9 | 99.9 | 99.9 | 100 | | 66.2 | 71.4 | 72.7 | 74.5 | 76.5 | 99.9 |
| bioS multi2 + permute | 100 | 100 | 100 | 100 | 100 | 100 | 99.9 | 100 | 100 | 100 | 100 | 100 | 99.5 | 99.7 | 99.8 | 99.9 | 100 | 100 | 99.9 | 100 | 100 | 100 | 100 | 100 | 93.3 | 95.3 | 96.8 | 98.0 | 98.8 | 99.9 | 90.2 | 92.8 | 95.0 | 96.8 | 98.6 | 100 |
| bioS multi2 + permute + fullname | 99.9 | 100 | 100 | 100 | 100 | 100 | 100 | 100 | 100 | 100 | 100 | 100 | 100 | 99.9 | 100 | 100 | 100 | 100 | 99.9 | 100 | 100 | 100 | 100 | 100 | 99.7 | 99.8 | 99.9 | 99.9 | 100 | 100 | 99.0 | 99.3 | 99.3 | 99.5 | 99.8 | 100 |
| bioS multi5 | 100 | | | | | | 44.6 | 100 | | | | | 44.0 | 77.2 | 99.8 | | | | 42.0 | 60.1 | 76.1 | 99.5 | | | 5.5 | 7.1 | 10.7 | 37.5 | 98.5 | | 14.0 | 13.8 | 14.8 | 21.6 | 56.2 | 100 |
| bioS multi5 + fullname | 100 | | | | | | 100 | 100 | | | | | 98.7 | 99.8 | 100 | | | | 99.3 | 99.9 | 99.9 | 99.9 | | | 98.1 | 99.6 | 99.7 | 99.7 | 100 | | 58.8 | 65.1 | 67.2 | 68.6 | 72.0 | 99.9 |
| bioS multi5 + permute | 100 | 100 | 100 | 100 | 100 | 100 | 100 | 100 | 100 | 100 | 100 | 100 | 100 | 100 | 100 | 100 | 100 | 100 | 100 | 100 | 100 | 100 | 100 | 100 | 99.9 | 99.9 | 99.9 | 100 | 100 | 100 | 99.9 | 100 | 100 | 100 | 100 | 100 |
| bioS multi5 + permute + fullname | 100 | 100 | 100 | 100 | 100 | 100 | 100 | 99.9 | 99.9 | 100 | 100 | 100 | 100 | 100 | 100 | 100 | 100 | 100 | 100 | 100 | 100 | 100 | 100 | 100 | 99.9 | 100 | 100 | 99.9 | 100 | 100 | 99.7 | 99.8 | 99.8 | 99.9 | 99.9 | 100 |

Figure 5: P-probing accuracies for various pretrained models on bioS data. Each **row** represents a pretrained model using a different knowledge augmentation, and each **column** labeled "$i$-$field$" shows the accuracy of predicting the *first token* of *field* from position $i$. Details are in Section 5 and Appendix E (where we also include experiments for the bioR data and for predicting the full-attribute *field*.) Details are in Section 5.1 and Appendix E (where we also include experiments for the bioR data and for predicting the full-attribute *field* in Figure 13 and 14.)

- For intermediate setups, the results are mixed. For example, comparing bioS single with multi5, we see that adding multiplicity (without permutation) results in earlier attribute storage, accounting for the increase in QA finetune accuracy from 9.7% to 41% as seen in Figure 3. Comparing bioS single+permute1 with single+permute5, we observe that permuting the six sentences five times (without diversifying the sentences) also leads to earlier knowledge storage, explaining the rise in QA finetune accuracy from 4.4% to 70%.

In sum,

> **Result 4** (Figure 5). *Increased knowledge augmentation in the pretrain data improves P-probing accuracies at earlier token positions. Consequently, a key-value pair knowledge (e.g., person-employer) more directly associates the value with the key rather than with other related attributes. This mechanism facilitates the (out-of-distribution) extraction of knowledge through fine-tuning.*

In Section 5, we use a Venn diagram to clearly demonstrate which attribute is stored upon observing another, further supporting this finding.

### 5.1.1 Closer P-Probing at Knowledge Dependency

As noted above, the model may infer attribute relationships based on their order in the pretrain data. For instance, if a birth date always precedes a company city, the model might infer "the person born on October 2, 1996 works in Menlo Park" instead of "Anya's work city is Menlo Park". This can occur if the pretrain data isn't adequately augmented, and the model may even favor linking one attribute to another, rather than to the person's name, if two attributes are closely correlated (such as company city and company name).

To investigate this, we created a variant of the bioS dataset, grouping the 6 sentences into 3 pairs with a consistent order: birthdate before birth city, university before major, and work company before work city. We allowed random permutations among these pairs and sentence diversities. We refer to this dataset as bioS couple (see Appendix A for details).

We examined our P-probing on this dataset as $2^5 \times 6$ classification tasks, predicting each of the six target attributes from a special token position where only a subset $S$ of the remaining five

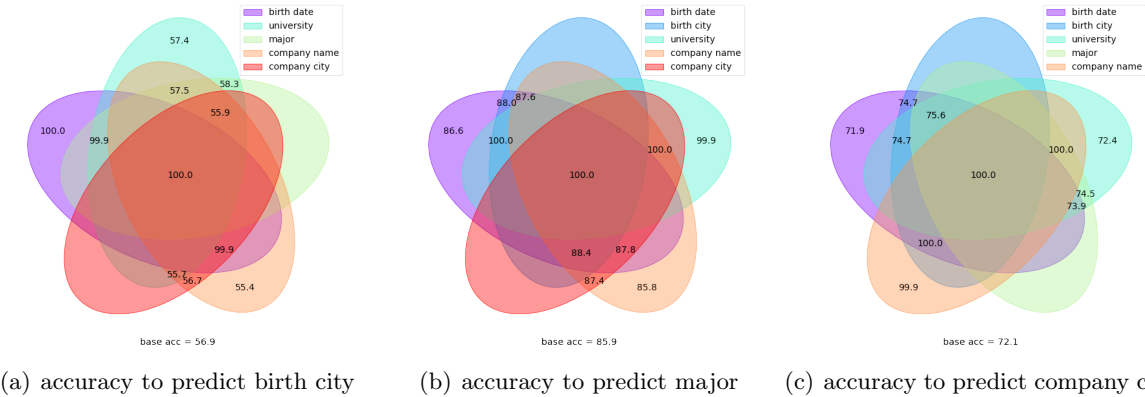

(a) accuracy to predict birth city    (b) accuracy to predict major    (c) accuracy to predict company city

Figure 6: Closer P-probing on bioS couple data in Section 5.1.1. The Venn diagram shows prediction accuracy for the target attribute at those special token positions, based on whether each of the remaining five attributes has been seen or not. More experiments like this are given in Figure 15 on Page 26.

> **Observation:** In this data the six attributes are coupled: birth date (resp. university, company name) always appears before birth city (resp. major, company city). We see significant accuracy improvement predicting birth city (resp. major, company city) after seeing birth date (resp. university, company name) attributes has been observed ($S$ has $2^5$ possibilities).[17] Our results, visualized in Figure 6, show that the accuracy in predicting the second attribute in each pair is heavily influenced by whether the model has encountered the first attribute, even with moderate data diversity.

*Remark* 5.1. This relates to Figure 3, where the "company city" attribute shows the weakest QA finetune performance in the bioS data family. This is due to our data construction, where "company city" is determined solely by "company name". The model thus associates "company city" with "company name" rather than the person's name, if the company name is presented earlier.[18]

### 5.1.2 P-Probing Extensions

We could consider alternative P-probing forms, such as introducing a low-rank update to the pretrained model's main body, like a trainable LoRA update with a small rank on the query/value matrices. While not necessary for our positive results (e.g., the highly augmented data bioS multi5+permute), it could be interesting to apply this to the negative results (e.g., the basic data bioS single). However, our experiments showed no significant increase in P-probing accuracies, so we omit the details.

Our P-probing has focused on the six distinct token positions, likely the preposition words preceding the six attributes. How about probing other positions, like tokens following each attribute or the person's name? We observed that P-probing accuracy might improve as the model processes more "extraneous" tokens. For instance, the P-probing accuracy for a person's birth date could increase after encountering phrases like "was born on" or "has birthday in". This could be due to the model's ability to associate the birthdate information with the sentence's *structure*. We chose not to include these observations for clarity.

---

[17]The P-probing process remains the same as before, using only 6 sets of trainable parameters each for a target attribute, each with a single classification linear layer and a single rank-2 update on the embedding. The difference is a more detailed interpretation of the results.

[18]This dynamic can be explored as a form of *knowledge manipulation*. For example, if the language model is good at retaining work company names, can it determine work city locations as a simple classification task using just company names? We explore this in our parallel paper [2].

Figure 7: Q-probing accuracies. Each **row** denotes a pretrained model with its specific knowledge augmentation. The left block reiterates QA finetune accuracies from Figure 3. The middle showcases Q-probing accuracies on the first-token prediction for the six attributes, and the right focuses on Q-probing for the whole-attribute prediction. (Further details for bioR and more are in Appendix E. Note: For birth date, first token predicts the whole birth month; we do not have whole-attribute prediction for it since it has too many choices.)

In Appendix E, we demonstrate the difference between a rank-2 and a rank-4 update on the embedding layer. It confirms that a rank-2 update is sufficient for P-probing on our datasets.

## 5.2 Result 5: Query-Based Probing

P-probing offers a qualitative assessment of early knowledge storage in the model relative to the original biography entry. However, it can be limiting due to its dependence on the exact context structure from the biography entry. For instance, in Figure 4, knowledge may be stored in short phrases like "received mentorship and guidance."

In query-based probing (Q-probing), we aim for a more precise, context-free value from a pretrained model, focusing on the knowledge directly associated with a person's name. We evaluate sentences containing only the person's full name and train a linear classifier on the last layer's hidden states to predict the person's six attributes. High accuracy suggests that the model directly links each person's attributes to their name.

We consider an input sentence containing only the person's full name, preceded by a starting token and followed by an ending token. Like P-probing, we freeze all transformer layers (acquired through pretraining), except the embedding layer, where we apply a low-rank update (using rank 16, compared to rank 2 in P-probing). This minimal change is necessary as we are addressing a distinct classification task under a different input distribution. We extract the hidden states from the last layer on the ending token and place a trainable linear classifier on top to predict the person's six attributes. More details are in Appendix F.

**Our findings.** Our results are in Figure 7. Our main finding is:

> **Result 5** (Figure 7). *The QA finetune accuracy* **correlates closely** *with Q-probing accuracy, indicating that the "degree to which the attribute is directly linked to the person's name" is a* **crucial factor for effective knowledge extraction.** *If the model fails to store knowledge in this way during pretraining, QA finetuning may not rectify this, regardless of the prompts or finetune parameters.*

Note, once again, applying knowledge augmentations to the pretrain data, Q-probing accuracy significantly increases. This suggests that the model encodes knowledge almost linearly in the hidden states directly adjacent to the person's name. Thus, the linear probes can extract the person's attributes from these hidden states as effectively as the model can be adapted through QA finetuning to answer questions related to those attributes.

| | QA mean acc | QA b_date | QA b_city | QA univ | QA major | QA c_name | QA c_city |
|---|---|---|---|---|---|---|---|
| baseline | 2.7 | 0.0 | 0.5 | 0.3 | 1.0 | 0.4 | 13.7 |
| bioS single + permute1 | 4.4 | 0.5 | 3.3 | 2.4 | 5.0 | 3.5 | 13.7 |
| bioS single + permute1 + CEL | 86.8 | 98.3 | 96.8 | 90.7 | 90.2 | 71.7 | 80.1 |
| bioR single | 10.0 | 25.1 | 13.9 | 2.4 | 5.5 | 2.0 | 14.1 |
| bioR single + wiki | 7.3 | 18.4 | 5.2 | 2.6 | 4.3 | 1.8 | 14.1 |
| bioR single + CEL | 76.3 | 94.3 | 85.3 | 82.9 | 79.4 | 67.0 | 56.6 |

Figure 8: QA finetune accuracy on the *minority group* with vs. without celebrity data in the pretraining process. Experiment details are in Appendix G, where we also include additional experiments in Figure 17.

Our result also suggests that, at the last hidden-layer, the model neither uses complex or non-linear transformations nor leverages interactions between hidden states at different token positions to extract knowledge about the person. This implies that the model **does not use contextual or global information from the biographies to extract knowledge about the individual**.

## 6   Result 6: Celebrity Can Help Minority

Section 4 highlighted the significant benefits of knowledge augmentation. However, in practice, we may not have augmented data for all individuals. This section explores whether partially augmenting data can improve knowledge extraction for non-augmented data. In our biography dataset, the augmented subset is akin to a "celebrity" group with plentiful online biographical information, potentially included in the fine-tuning dataset as well. The non-augmented subset is comparable to a "minority" group with limited biographical data.

For comparison, we introduce an additional set of $N = 100,000$ individuals, the celebrity group $\mathcal{P}_{\text{cel}}$, while the original $N$ individuals form the minority group $\mathcal{P}_{\text{min}}$. We test both synthetic bioS and more realistic bioR data. For bioS, the celebrity group's biographies use the multi5+permute augmentation, simulating varied expressions found on internet. For bioR, the celebrity group uses the multi5 augmentation, generating their biographies five times using Llama.

The language model is pretrained on the combined set $\mathcal{P}_{\text{cel}} \cup \mathcal{P}_{\text{min}}$ biographies and then fine-tuned using QAs from the celebrity group $\mathcal{P}_{\text{cel}}$. We evaluate the model's QA accuracy on the $\mathcal{P}_{\text{min}}$ group.[19] Our results are presented in Figure 8.

> **Result 6** (Figure 8). *Introducing celebrity data boosts the minority group's QA accuracy (e.g., from 4.4% to 86.8% for the bioS data). This is significant because:*
> - *the minority group's BIO pretrain data remains unchanged in both cases, and*
> - *the minority group's QA data is not used during fine-tuning.*

This highlights that **merely including celebrity data during pretraining** significantly improves the model's ability to store and extract knowledge from the minority group. Similarly, in the more realistic bioR case, introducing celebrity data increases the minority's accuracy from 10.0% to 76.3%. This strongly suggests that this phenomenon *also occurs in real-world scenarios*.

We also use P-probing and Q-probing techniques to validate and explain the above findings; they suggest that with the inclusion of celebrity data, the attributes of the minority group are more directly stored onto their names. These are detailed in Figure 18 and Figure 19 in Appendix G.

---

[19] Other fine-tuning variations, such as QA fine-tuning with half of $\mathcal{P}_{\text{min}}$ as training and half as testing, show negligible differences.

Figure 9: Additional results on the GBERT model pretrained via masked language modeling (MLM). Mixed training (left) versus BIO pretrain + QA finetune (middle left) versus Q-probing (middle right and right).

> **Observation.** MLM doesn't necessarily promote proper knowledge storage for subsequent extraction; unless the knowledge is a single word or consists of independent words (like month, day, year), extracting knowledge after MLM pretraining might still be nearly impossible. (Further details are in Appendix H, and note we have pretrained trained the model twice longer comparing to GPT.)

*Remark* 6.1. The benefit of celebrity data is *not universal*. Substituting it with the WikiBook dataset improves the model's English comprehension, yet it still struggles with biographical knowledge extraction. This suggests that only celebrity data of *similar form* truly aids knowledge extraction for minority groups. In Figure 17 in Appendix G, we further investigate different celebrity data types and instances of minor format differences between minority and celebrity knowledge.

## 7 Result 7: Knowledge Storage for Bidirectional Models

This paper primarily explores knowledge storage and extraction in auto-regressive language models. One may argue that some knowledge issues, such as the consistent knowledge ordering in bioS single, are unique to this task due to its *unidirectional* nature. We thus pose the question, *Could BERT be a solution to this?*

We analyze the BERT model [20], similar to GPT2 but with a full attention matrix, allowing every token to attend to every other token. For a direct comparison, we modify our GPT2 architecture to replace its triangular attention matrix with a full matrix, keeping the GPT2 tokenizer and rotary embedding. We call this modified model GBERT.

Our pretraining task is now *whole-word masked-language modeling (MLM)*. Each English whole-word has a 15% chance of being selected, which is then replaced with a <MASK> token (80% chance), retained (10% chance), or replaced with a random token (10%). The goal is to predict the original word for these selected tokens.[20]

For GBERT, we modify the QA task to evaluate its knowledge extraction capabilities. For questions like "What is the birth city of Anya Briar Forger?", we append them with several <MASK> tokens (equaling the answer's length).[21] A correct answer requires accurate recovery of all masked tokens.

We display results for both mixed training and BIO pretrain + QA finetune. Half of the QAs are used for mixed training (or QA fine-tuning), while we test out-of-distribution generalization accuracies on QAs for the remaining half of the people. Q-probing results for GBERT are also presented, determining if the model, with minor embedding layer modifications, can linearly predict target attributes from a person's full name.

---

[20]We thank Xiaodong Liu and Pengcheng He from the `mt-dnn` project [24] for confirming that our MLM implementation aligns with common practice.

[21]Revealing the answer's token count might seem unfair. However, given our aim to highlight GBERT's limitations, this extra information doesn't hinder our intentions.

**Our findings.** Our findings are displayed in Figure 9. Key observations include:

- The QA-finetune and Q-probing accuracies again show a strong correlation. This suggests that the ability to extract knowledge from a BERT-like model *also depends* on whether such information is nearly linearly stored in hidden states directly adjacent to the person's name. This means our Q-probing technique is effective also for encoder models like BERT.

- Consistent with Figure 3, mixed training yields slightly superior out-of-distribution QA accuracies compared to BIO pretrain + QA finetune.

- Interestingly, the model performs well on "birth date" and "major" attributes but struggles on others. The reason is simple. In MLM, where each word has an equal chance of being masked, the model learns to associate knowledge words with the *most related* unmasked word, preferably those that are *adjacent.* For instance, words representing the "birth date" attribute (month, day, year) are quite independent, making the model more inclined to link them to the person's name. For attributes like birth city, where there's a strong link between the city "Bellevue" and state "WA", the model maximizes this association, *inhibiting storage of knowledge on person names.*[22]

**Result 7** (Figure 9). *While bidirectional models like BERT is less sensitive to knowledge ordering, the MLM pretraining task **does not** necessarily promote knowledge storage for subsequent extraction. **Unless the knowledge is a standalone word or of independent words** (like month, day, year), extracting knowledge after MLM pretraining might prove challenging, if not totally impossible.*

## 8 Conclusion

This study explores the capability of pre-trained language models to store and retrieve knowledge through question-answering tasks. We developed a controlled biography dataset and employed probing techniques to assess how knowledge augmentation influences the extractability of knowledge in pre-trained transformer models. Using synthetic data allows for greater control over the training and fine-tuning of models, which is essential for understanding how different data sources impact the **internal mechanisms** of transformers. This could be a significant future direction for unraveling the complexities of transformers.

For practitioners, this paper emphasizes the **importance of rewriting** critical but infrequent data **during the pretrain stage** to enhance knowledge extraction for downstream tasks. Tools like ChatGPT, Llama-7B, or smaller auxiliary models can be used for rewriting before pre-training; these models do not need to contain the knowledge themselves, and even simple techniques like sentence-level shuffling or English-to-French translation can be beneficial.

Additionally, we suggest **including more instruction-finetuned data** during the pretrain phase. Although this approach differs from human knowledge acquisition, it enhances the model's ability to encode knowledge more effectively, as explored in recent follow-up work [19].

Finally, Part 3 of this work series focuses on how language models store, extract and manipulate knowledge (including Part 3.2 [2] and Part 3.3 [3]). We also cover grade-school math and reasoning in Part 2 [38, 39], and learning hierarchial language structures in Part 1 [1].

---

[22]Similarly, many majors are single words so this explains its high QA test accuracy. In contrast, the words representing universities or company names/cities are more dependent.

# Appendix

## A Details on Data Preparation

### A.1 BIO dataset bioS

In the synthetic dataset labeled as bioS, we generate profiles for $N = 100,000$ individuals. Each individual's first, middle, and last names, birth date, birth city, university attended, major of study, and current employer are selected *independently* and randomly from a uniform distribution.

- First, middle, and last names are drawn from pools of 400, 400, and 1000 English names respectively. We apply rejection sampling to ensure all $N$ individuals have unique full names.
- Birth years range from 1900 to 2099, months are selected from the 12 months, and days are chosen between 1 and 28.
- Birth cities are selected from 200 US cities, with their respective state abbreviations, such as `Princeton, NJ` and `Cambridge, MA`.
- Universities are drawn from a list of 300 US institutions. Some may have similar prefixes, like `University of California, Berkeley/Irvine/Davis/etc`.
- Majors are selected from 100 common college disciplines, including Computer Science, Physics, and Music.
- Employers are chosen from a list of 263 companies, featuring names like Meta Platforms, Microsoft, and Google.

Additionally,

- We introduce a "company city" attribute that ***depends*** on the US location of the employer's headquarters. For instance, an employee of Meta would list Menlo Park, CA as their company city. Notably, 13.7% of the companies are headquartered in New York, NY. Thus, defaulting to New York, NY when predicting a person's work city yields a base accuracy of 13.7%.

In the bioS dataset, we craft a biographical text entry for each individual, distilling their profile into six sentences. Each sentence illuminates a distinct attribute of the individual. To increase diversity, we select each sentence randomly from a set of pre-defined templates. Specifically, we have 46 sentence templates for birth dates, 49 for birth cities, 49 for universities, 52 for majors of study, 47 for employers, and 48 for company cities. Beyond (2.1), we provide several more examples below:

Carlos Jameson Stokes has his annual celebration on November 12, 2088. He celebrates his birth in San Francisco, CA. He graduated from Oklahoma State University. He explored the theoretical aspects of Information Systems. He contributed his expertise to United Airlines Holdings. He acquired industry knowledge while working in Chicago, IL.

Alondra Bennett Rooney celebrates their life journey every year on April 1, 1909. They owe their roots to Durham, NC. They benefited from the resources and facilities provided by University of South Alabama. They developed a strong foundation in Data Science. They had a job at The Southern Company. They were involved in the industry of Atlanta, GA.

Aidan Alexa Dennis's birth is celebrated annually on July 17, 1968. She calls Palmdale, CA her birthplace. She specialized in her field of study at Stevens Institute of Technology. She completed a rigorous program in International Business. She had employment prospects at Johnson & Johnson. She gained work experience in New Brunswick, NJ.

(We assign a random pronoun (he/she/they) to each person.)[23]

---

[23] Given that we are not employing a pretrained model sourced from the internet, we did not do fact-checking. For instance, a person's major may not align with the business of the company they work for, and their birth year might largely precede the company's establishment date.

In the basic configuration, we produce *a single biographical entry* for each individual, maintaining a consistent order for the six sentences as previously outlined. In average, a biographical entry has 73.0 tokens using GPT2 tokenization. We denote this configuration as "bioS single." For comparison, we delve into 15 knowledge augmentations:

- bioS single+fullname: Pronouns are replaced with the person's full name.
- bioS single+permute1/2/5: The six sentences in the biography entry are randomly permuted 1/2/5 times for each person. However, the full name only appears in the first sentence, with subsequent sentences using pronouns. This results in 1/2/5 biography entries for each person.
- bioS single+permute1/2/5+fullname: As with the previous augmentation, but the full name is used in all six sentences.
- bioS multi2/5: 2 or 5 biographical entries are generated for each person, with each generation employing a re-sampled set of sentence templates.
- bioS multi2/5+permute: Building on bioS multi2/5, the six sentences within each biographical entry are randomly permuted. However, the full name appears only once in the first sentence.
- bioS multi2/5+fullname: Building on bioS multi2/5, pronouns are replaced with the individual's full name across all sentences.
- bioS multi2/5+permute+fullname: Incorporating features from both bioS multi2/5+permute and bioS multi2/5+fullname, the pronouns are replaced with the individual's full name and the six sentences are randomly permuted.

### A.1.1  bioS couple

In Section 5.1.1, when delving deeper into P-probing, we also introduced a partial knowledge augmentation on the bioS dataset, which we termed bioS couple.

Specifically, we initially generate six sentences, each derived from a set of sentence templates similar to those in bioS single. We then group these six sentences into three pairs. The sentence describing a person's birthdate always precedes the one discussing the person's birth city. Similarly, the sentence detailing the person's university consistently comes before the one about their major, and the one about their employer invariably precedes the sentence regarding their work city. Subsequently, we permute the order of these three pairs of sentences, resulting in $3! = 6$ potential arrangements. The individual's full name is restricted to appear only in the first sentence. For each individual, we create such a biographical entry 1/2/5 times, designating this dataset as bioS couple1/couple2/couple5. Our experiments in Figure 6 were with respect to bioS couple2, and we shall give the similar results in Figure 15 for bioS couple1/5 for comparison.

### A.2  BIO dataset bioR

We examine a "close-to-real" dataset produced by Llama [37, 40]. Specifically, for the previously set of $N = 100,000$ individuals, we provide an instructive prompt to Llama1-30B, directing it to generate a biographical entry for us.[24] To promote diversity in our results, we draw our instructive prompts from a collection of sentence templates. Three examples are presented below.

> Llama prompt: Write a 50 words biography about a person whose name is Nicole Kevin Pratt. The person studied at Haverford College. The person studied Management there. The person was born and raised in Baltimore, MD. The person worked in Cincinnati, OH for Procter & Gamble. The person's birthday falls on January 25, 1977.

---

[24] Paraphrasing is simple, so larger models are unnecessary as results are similar (our trials confirmed this, but including them seemed uninteresting). At the time of paper writing, Llama2 was not released (but subsequent tests using Llama2 yield similar results, see [3]).

Llama prompt: Write a 50 words biography about a person whose name is Hunter Bennett Kenny. The person attended Queens College, City University of New York for education. The person pursued a degree in Political Science there. The person originated from Augusta, GA. The person worked in Menomonee Falls, WI for Kohl's. The person was born on March 25, 2033.

Llama prompt: Write a 50 words biography about a person whose name is Johnathan Charles Wade. The person attended Colorado State University for education. The person pursued a degree in Sociology there. The person was brought up in New York City, NY. The person worked in Northbrook, IL for Allstate. The person's birthday falls on January 7, 2098.

Here are the generated outputs from Llama [37, 40]:

Nicole Kevin Pratt is an American business executive. She is currently the Vice President of P &G Global Business Services at Procter & Gamble. She was born on January 25, 1977, in Baltimore, Maryland. She graduated from Haverford College with a degree in Management. P &G recruited her as an Assistant Brand Manager in 2000. She held various leadership positions in brand management, marketing, and sales across different business units and categories. She was named Vice President of P &G Global Business Services in 2019. Nicole currently lives in Cincinnati, Ohio with her husband and three children.

Hunter Bennett Kenny is a talented political science graduate from Queens College, City University of New York. He hails from Augusta, Georgia and was born on March 25, 2033. During his time at college, he was an active member of the student council and served as its president in his senior year. He interned at the office of New York Senator Chuck Schumer. After graduating cum laude, he worked for Kohl's in Menomonee Falls, Wisconsin. He currently resides in Brooklyn, New York.

Johnathan Charles Wade is a successful insurance agent who works for Allstate. He was born on January 7, 2098, in New York City, NY. He graduated from Colorado State University, where he majored in Sociology. He currently resides in Northbrook, IL.

We observed that Llama might not always generate a biographical entry that includes all six attributes. To address this, we repeat the sampling process until Llama's output ensures the inclusion of all attributes. Typically, the entry begins with the individual's full name, and the oder in which the six attributes appear can vary. In average, a biographical entry has 72.3 tokens using GPT2 tokenization.

In the basic configuration, we produce a single biographical entry for each person, denoted as "bioR single." For comparison, we also introduce the multi$M$ augmentation, which creates $M$ entries per person, and the fullname augmentation.

## B  Details on Model Architecture

The GPT2-small architecture [31] has 12 layers, 12 heads, and $768 = 12 \times 64$ hidden dimensions (124M). Recent research [8, 16, 34] has shown that transformers can achieve a significant performance improvement by utilizing attentions based on the *relative* positional differences of tokens. (This is more systematically studied in [1].) Consequently, in this paper, we replace the positional embedding with a rotary embedding, following the standard GPT-NeoX implementation [8] available on Huggingface (with the default frequency base of 10,000 and rotary dimension set to a 1/4 of the embedding dimension). We continue to refer to this as GPT2 for brevity.

In our bioS experiments, we employ the above architecture. For the bioR experiments, we opt for a larger GPT2 model with 12 layers, 20 attention heads each 64-dimensional (302M), tailored to its increased difficulty. Only when presenting our negative result in Figure 2, we also tried a 12-layer, 32-head (each 64-dimensional) GPT2 model (682M).

In our knowledge extraction experiments (e.g., Figure 3), we also used a downsized Llama architecture with the same number of layers, heads, and hidden dimensions as the GPT2 architecture, yielding similar results. For the probing experiments, we exclusively used the GPT2 architecture for conciseness.

Additionally, we evaluate the BERT model [20]. BERT is similar to GPT2 but features a complete attention matrix, enabling every token to attend to all others. For a strong side-by-side comparison, we modify our GPT2 architecture to swap its triangular attention matrix for a full matrix, while keeping the GPT2 tokenizer and rotary embedding (removing positional embedding).

We label this revised model GBERT. A primary distinction is that GBERT adopts pre-layernorm (inherited from the base GPT2 architecture), whereas BERT utilizes post-layernorm.

Throughout pretraining, mixed training, and QA finetuning, we maintain a context window length of 512.

## C  Details on Pretrain and Mixed Training

| | 0.1 | 0.2 | 0.3 | 0.4 | 0.6 | 0.8 | 0.9 |
|---|---|---|---|---|---|---|---|
| bioS single | 24.5 | 53.6 | 70.5 | 77.5 | 87.1 | 85.7 | 81.8 |
| bioS single + permute1 | 11.7 | 25.5 | 55.1 | 69.3 | 85.1 | 82.4 | 83.8 |
| bioS single + permute2 | 87.1 | 91.4 | 91.5 | 90.1 | 93.2 | 91.6 | 84.6 |
| bioS multi2 + permute | 94.3 | 95.1 | 95.5 | 95.3 | 95.3 | 91.3 | 91.9 |
| bioS multi5 + permute | 96.6 | 96.3 | 95.9 | 96.2 | 95.8 | 95.2 | 93.9 |

QA$_r$ - ratio of QA data in mix training

(a) bioS

| | 0.1 | 0.2 | 0.3 | 0.4 | 0.6 | 0.8 | 0.9 |
|---|---|---|---|---|---|---|---|
| bioR single | 14.7 | 22.4 | 28.3 | 35.1 | 71.4 | 78.7 | 78.4 |
| bioR single + fullname | 63.8 | 70.2 | 75.2 | 78.3 | 80.3 | 81.9 | 80.1 |
| bioR single + permute5 | 58.3 | 68.2 | 74.3 | 81.5 | 82.9 | 86.6 | 84.5 |
| bioR multi3 | 93.8 | 93.9 | 94.0 | 93.4 | 93.4 | 93.0 | 91.5 |
| bioR multi5 | 94.3 | 94.0 | 93.9 | 94.7 | 94.5 | 93.7 | 93.2 |

QA$_r$ - ratio of QA data in mix training

(b) bioR

Figure 10: QA test accuracy for mixed training across various choices of QA$_r$.

> **Observation:** Particularly with more challenging data (i.e., with less knowledge augmentation), a greater QA ratio during training results in enhanced out-of-distribution QA (test) accuracy. This lends further credence to our assertion about the model's unusual behavior: it initially acquires knowledge from QA rather than BIO and subsequently seeks to link BIO with QA.

During BIO pretraining, we randomly sample biographical entries of individuals and concatenate them to form sequences of 512 tokens, using a standard `<EOS>` token to separate individual entries.

In mixed training, we pre-train the model with BIO data from *all* individuals and QA data from *half* of them. Specifically, each training sequence of 512 tokens is either sourced entirely from the BIO entries (as previously mentioned) or entirely from the QA entries (again, from randomly sampled individuals and concatenated). We define a parameter QA$_r$ to dictate the frequency of using QA entries. Predominantly in this paper, we set QA$_r$ = 0.8, which implies a 2 : 8 ratio between BIO and QA entries in terms of the number of pre-trained tokens. We subsequently assess the model's generation accuracy using QA data from the other half of the individuals. Refer to Figure 10 for an analysis of how the parameter QA$_r$ impacts mix-training performance.

For both BIO pretraining and mixed training, we employed a conventional set of optimization parameters: the AdamW optimizer with a weight decay of 0.1, $\varepsilon = 10^{-6}$, an initial learning rate of 0.001, a 1000-step linear warmup, and cosine learning rate decay (from 0.001 decreasing to 0.0001). We used a batch size of 96.

There were a total of 80,000 training steps for bioS (using the 12-layer, 12-head GPT2/Llama architecture) and 150,000 training steps for bioR (using the larger 12-layer, 20-head GPT2/Llama). Only when using the 12-layer, 32-head GPT2 to present our negative result in Figure 2, we used 200,000 training steps.

*Remark* C.1. Our training time is long enough to ensure next-token prediction accuracy well above 99% for both BIO pretraining and mixed training, when focusing on tokens describing six attributes per individual. These numbers are not included in this paper's figures.

## D  Details on QA Finetune

In our QA finetuning tasks, we first use a BIO pretrained model checkpoint and then apply either full finetuning or LoRA finetuning.

Figure 11: BIO pretrain + QA finetune (train acc) / **test acc** for various choices of fine-tuning settings. Bold number indicates QA generation accuracy on $\mathcal{P}_{\text{test}}$, and the smaller number in bracket represents QA (first-token) accuracy on $\mathcal{P}_{\text{train}}$. For LoRA fine-tune we consider a rank $r = 2, 4, 8, 16, 32$ update on the query/value (q/v) matrices and a rank $r' = 0, 16, 32, 64, 128$ update on the word embedding matrix. This is an extension of Figure 2.

(a) Analogous to Figure 3 but for the bioR data family using the GPT2 architecture

(b) Analogous to Figure 3 but for the bioR data family using the Llama architecture

(c) Analogous to Figure 3 but for the bioS data family using the Llama architecture

Figure 12: Comparison of BIO pretraining + QA finetuning (**left**) versus their Mixed Training counterparts (**right**) under various knowledge augmentations on the data (the **rows**).

For full finetuning, we employ the AdamW optimizer with $\varepsilon = 10^{-6}$. We use weight decays of 0.01 and 0.001, and initial learning rates of 0.001, 0.0003, and 0.0001. There is no warmup, and we implement cosine learning rate scheduling (reducing to 10% of the initial learning rate), a batch size of 48, and a total of 50,000 training steps. Given that we are presenting a negative result for full finetuning (as seen in Figure 2), we display the best QA test accuracy among all the lr/wd parameter combinations.

For LoRA finetuning, we maintain the aforementioned AdamW configuration but set a consistent weight decay of 0.01 and an initial learning rate of 0.0003 for all tasks.

The results in Figure 11 suggest that for the purpose of QA finetuning, LoRA is generally a better option compared to full finetuning. While a large rank-$r$ update on the query/value matrices isn't essential, it appears beneficial to have a significant rank-$r'$ update on the embedding layer to address the distribution shift from the BIO data to the QA data.

For this reason, in all subsequent experiments in this paper (notably Figure 3 and 12), when conducting QA finetuning, we use $r' = 128$ and either $r = 8$ or $r = 16$, presenting the best accuracy from the two runs.

# E   Details on P-probing

In our P-probing experiments, we freeze the BIO pretrained GPT2 model and append a limited set of trainable parameters. Using the GPT2-small as an example, we introduce:

- a trainable rank-2 update for the embedding layer, having dimensions of $50256 \times 2$ and $2 \times 768$,
- for each prediction task that is an $M$-class classification problem, a trainable linear layer with dimensions of $768 \times M$,
- preceding the linear layer, a layer normalization layer furnished with trainable affine parameters.

In the context of P-probing, recall that we considered six classification sub-tasks (from 6 special locations) for every attribute prediction task. Specifically, for the birthdate attribute, we solely address its first-token prediction task, which is equivalent to predicting the individual's birth month.[25] For the remaining five attributes, both the first-token and whole-attribute prediction tasks are examined. In sum, this results in 11 prediction tasks, each comprising 6 sub-tasks. For every one of these 11 tasks, we incorporate a distinct set of trainable parameters.

For optimization, the AdamW optimizer is employed with $\varepsilon = 10^{-6}$, weight decay of 0.3, an initial learning rate of 0.001, no warmup, and a linear learning rate decay (down to 0 in the end). We set the batch size of 50 and trained for 30,000 steps. During this P-probing training phase, we have turned on the dropout on the (frozen) pretrained GPT2 model to prevent overfitting.

We perform experiments on both bioS and bioR data families (refer to Figure 13 and Figure 14), evaluating the P-probing accuracy of first-token and whole-attribute predictions. These figures also compare rank-2 and rank-4 updates on the embedding layer, demonstrating that a large modification to this layer is not crucial for P-probing attribute values.

---

[25]This is because a birthdate encompasses $200 \times 12 \times 28$ potential choices, surpassing $N/2$, the number of training individuals.

(a) P-probing first-token prediction accuracy; LoRA embedding layer rank = 2

(b) P-probing first-token prediction accuracy; LoRA embedding layer rank = 4

(c) P-probing whole-attribute prediction accuracy; LoRA embedding layer rank = 2

(d) P-probing whole-attribute prediction accuracy; LoRA embedding layer rank = 4

Figure 13: **P-probing accuracies on the bioS data (extension of Figure 5).** Each row represents a different pretrained model using its associated knowledge augmentation on the bioS data. For every $i \in \{0, 1, \ldots, 5\}$ and $field \in \{\text{bmonth,bcity},\ldots\}$, the column labeled "$i\text{-}field$" shows the accuracy when predicting the first token / whole attribute of $field$ from the special position $i$.

**Observation.** Comparison between LoRA rank 2 and 4 shows that a rank-2 update on the embedding layer is sufficient for P-probing purposes. The P-probing results for the whole-attribute scenario largely align, but when predicting longer attributes, like "university", the classification accuracy falls short of 100%. This outcome is consistent with expectations: extracting partial knowledge from subsequent tokens in a lengthy attribute can be difficult, as further detailed in our companion paper [2].

(a) P-probing first-token prediction accuracy; LoRA embedding layer rank = 2

(b) P-probing first-token prediction accuracy; LoRA embedding layer rank = 4

(c) P-probing whole-attribute prediction accuracy; LoRA embedding layer rank = 2

(d) P-probing whole-attribute prediction accuracy; LoRA embedding layer rank = 4

Figure 14: **P-probing accuracies on the bioR data (extension of Figure 5).** Each row represents a different pretrained model using its associated knowledge augmentation on the bioR data. For every $i \in \{0, 1, \ldots, 5\}$ and $field \in \{\text{bmonth,bcity,}\ldots\}$, the column labeled "$i$-$field$" shows the accuracy when predicting the first token / whole attribute of $field$ from the special position $i$.

**Observation.** P-probing results on the bioR data family closely mirror those on bioS. Incorporating additional knowledge augmentations in the pretrain data enables the P-probing accuracies to improve at earlier special positions.

## E.1 Details on Closer P-Probing

In Figure 6 and Section 5.1.1, we examined the P-probing results using a Venn diagram based on the bioS couple dataset from Appendix A.1.1, where each individual has 2 biographical entries. Figure 15 supplements this with results for individuals with 1 or 5 entries.

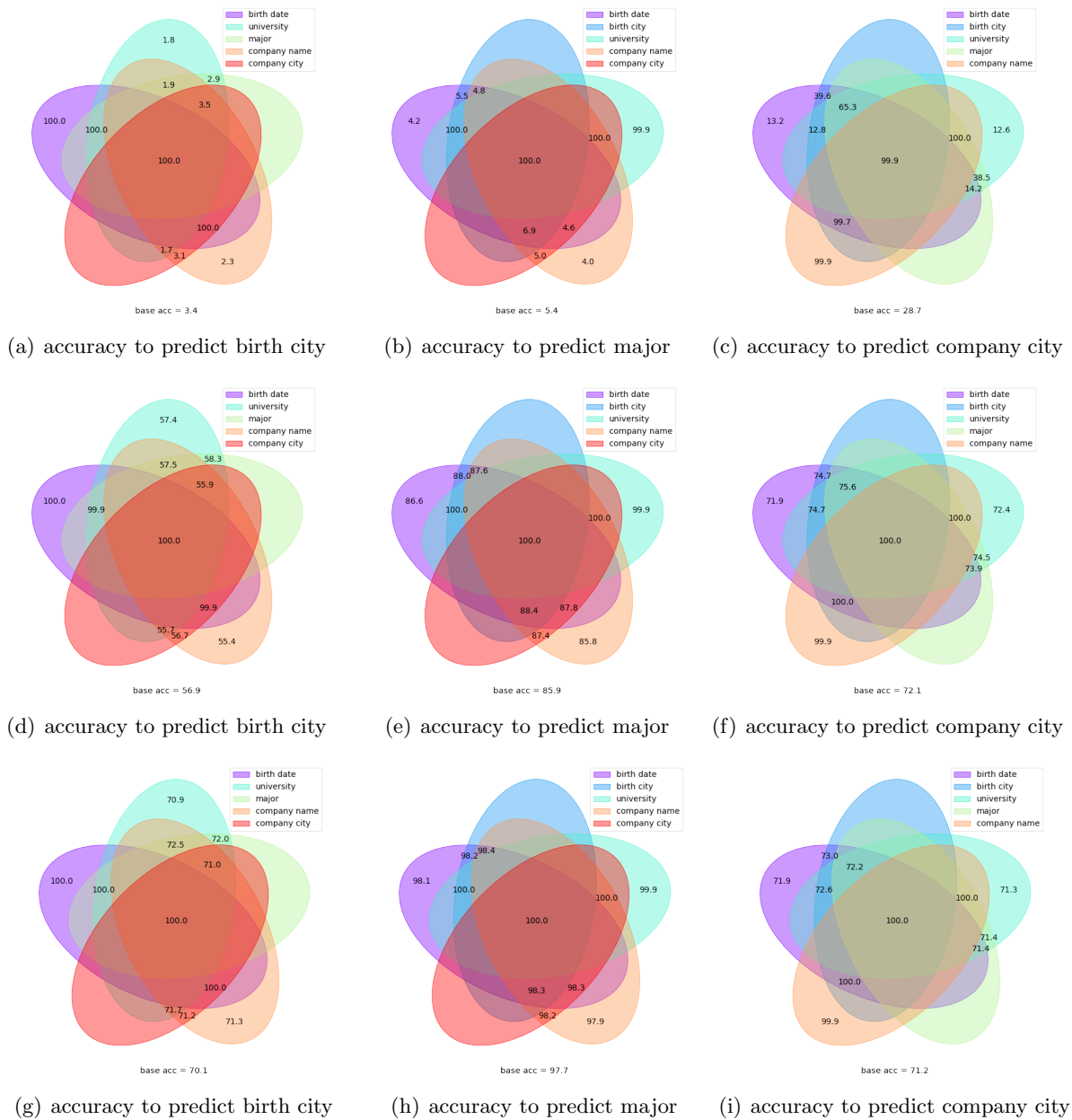

Figure 15: This is an extension of Figure 6 with more data: bioS couple1 (top), bioS couple2 (middle), and bioS couple5 (bottom). The Venn diagram shows prediction accuracy for the target attribute at those special token positions, based on whether each of the remaining five attributes has been seen or not.

**Observation:** Again, we see accuracy improvement predicting birth city (resp. major, company city) after seeing birth date (resp. university, company name) Such knowledge dependency can be somewhat mitigated with the introduction of more data diversity, but not in full.

# F  Details on Q-probing

Recall that in Q-probing, we freeze the pretrained GPT2 model and append a small set of trainable parameters on top for probing purposes. Using GPT2 small as an example, we add:

- a trainable rank-$r$ update on the embedding layer with dimensions of $50256 \times r$ and $r \times 768$,
- a trainable linear layer with dimensions of $768 \times M$ for each prediction task that is an $M$-class classification problem,
- a batch normalization layer before the linear layer, with trainable affine parameters.

We consider an input sentence that *only* contains a person's full name, preceded by a starting token and followed by an ending token. After applying all 12 layers of GPT2, we extract the hidden states from the last layer at the ending token. For instance, in the GPT2-small model, this is a 768-dimensional vector. We then apply a linear classifier on top to predict the person's attributes. Similar to P-probing, we adopt a separate set of trainable parameters for each of the 11 classification tasks.

We employ the AdamW optimizer with $\varepsilon = 10^{-6}$, a weight decay of 0.3, an initial learning rate of 0.001, no warmup, and a linear learning rate decay schedule (reducing to 0 by the end). The batch size is set to 200, and we run a total of 30,000 training steps. During training, we allow the frozen GPT2 model to use dropout.

Experiments are conducted on both the bioS and the bioR data families, as shown in Figure 16, for first-token prediction and whole-attribute prediction. We compare rank-16 versus rank-64 updates on the embedding layer for the bioS data (or rank-32 versus rank-128 updates for the bioR data). This demonstrates that for Q-probing, a larger modification to the embedding layer is not necessary to probe the desired attribute values.

(a) Q-probing for the bioS data family; LoRA embedding layer rank = 16

(b) Q-probing for the bioS data family; LoRA embedding layer rank = 64

(c) Q-probing for the bioR data family; LoRA embedding layer rank = 32

(d) Q-probing for the bioR data family; LoRA embedding layer rank = 128

Figure 16: Q-probing accuracies (extension of Figure 7). Each row denotes a pretrained model with its specific knowledge augmentation. The left block reiterates QA finetune accuracies from Figure 3 and Figure 12. The middle showcases Q-probing accuracies on the first-token prediction for the six attributes, and the right focuses on Q-probing for the "whole-attribute" prediction.

---

**Observation.** Comparison between LoRA ranks show that a rank-16 (resp. rank-32) update on the embedding layer is sufficient for Q-probing purposes on bioS (resp. bioR). Q-probing results on the bioR data family closely mirror those on bioS. Incorporating additional knowledge augmentations in the pretrain data enables the Q-probing accuracies to significantly improve.

## G  Details on Celebrity Augmentation

Recall that in the celebrity knowledge augmentation, we introduced an additional set of $N = 100,000$ individuals and designated them as the celebrity group, $\mathcal{P}_{\mathsf{cel}}$. In contrast, the original $N$ individuals represent the minority group, $\mathcal{P}_{\mathsf{min}}$. There is no overlap between these two sets of individuals; specifically, they have distinct full names.

In the main body of this paper (specifically in Figure 8), we considered two choices:

- The minority uses bioS single+permute1, and the celebrity uses bioS multi5+permute. We denote this combination as bioS single+permute1+CEL and compare it to bioS single+permute1.

- The minority uses bioR single, and the celebrity uses bioR multi5. We denote this combination as bioR single+CEL and compare it to bioR single.

  (We also compare the latter to bioR single+wiki. By this, we mean that during BIO pretraining, half of the training sentences come from the WikiBook dataset, while the other half come from the bioR single data.)[26]

Note that in both cases, each individual in the minority group has only one biographical entry, while each individual in the celebrity group has five biographical entries. Thus, during BIO pretraining, the BIO data on $\mathcal{P}_{\mathsf{cel}}$ appear with a 1/6 chance.

In this appendix, we explore a broader set of augmentation options.

- The minority uses bioS single and the celebrity uses bioS multi5+permute, denoted as bioS single+CEL. We compare this to bioS single. In this scenario, the celebrity and minority groups **have biographical entries in different formats**: the entries of the celebrity group are *randomly shuffled*, while those of the minority group follow a *fixed order* (see (2.1)). The QA test accuracy on the minority group increases with the addition of the celebrity group, but not to the same extent as in the bioS single+permute1+CEL case.

- The minority uses bioS single+permute1+fullname and the celebrity uses bioS multi5+permute, denoted as bioS single+permute1+fullname+CEL. We compare this to bioS single+permute1+fullname. In this scenario, the celebrity and minority groups **have their biographical entries in different formats**: the minority group uses the fullname augmentation, repeating the individual's full name in each sentence, while the celebrity group only mentions the fullname once. The QA test accuracy on the minority group increases with the assistance of the celebrity group, but not as much as in the bioS single+permute1+CEL case.

- The minority uses bioR single+fullname and the celebrity uses bioR multi5+fullname, denoted as bioR single+fullname+CEL. We compare this to bioR single+fullname. In this case, the celebrity and minority groups **have their biographical entries in the same format**, leading to a significant increase in QA test accuracy to 82.2%.

  (We also compare this to bioR single+fullname+wiki, where during BIO pretraining, half of the training sentences come from the WikiBook dataset, and the other half from the bioR single+fullname data. C.f. Remark 6.1)

The transformer model is pretrained on the combined set of biographies $\mathcal{P}_{\mathsf{cel}} \cup \mathcal{P}_{\mathsf{min}}$ and then finetuned using QAs from the celebrity group $\mathcal{P}_{\mathsf{cel}}$. We evaluate the model's QA generation accuracy on the $\mathcal{P}_{\mathsf{min}}$ group.[27] Our findings are reported in Figure 17.

---

[26] Recall that BERT and RoBERTa were trained on a combination of BookCorpus [41] and English Wikipedia, which totals 16GB of uncompressed text [20, 25]. We use this same 16GB WikiBook dataset.

[27] We also considered other fine-tuning variations, such as QA finetuning with half of $\mathcal{P}_{\mathsf{min}}$ as training and half as testing, but found negligible differences.

Figure 17: QA finetune accuracy on the *minority group* with versus without celebrity data in the pretraining process. This is an extension to Figure 8, and the details are given in Appendix G.

**Observation.** The augmentation effect from the celebrity data may be weakened if the minority group uses differently formatted BIO data, such as using full names when the celebrity does not (see bioS single+permute1+fullname+CEL), or maintaining a fixed sentence order when the celebrity does not (see bioS single+CEL). We also conducted an experiment where both the celebrity and minority used bioR data with full name augmentation. In all cases, incorporating celebrity data significantly improved QA test accuracy for the minority group.

**P-probing and Q-probing.** We incorporate P-probing and Q-probing results for our celebrity case. The inclusion of celebrity data enhances the model's structural knowledge storage, *even for minority groups*. Figure 18 demonstrates that knowledge about minority groups is often stored in earlier tokens. This confirms that for *minority groups*, individual full names can more directly encode the six target attributes, due to the introduction of celebrity data. This accounts for the high knowledge-extraction QA accuracies.

(a) P-probing first-token prediction accuracy; LoRA embedding layer rank = 2

(b) P-probing whole-attribute prediction accuracy; LoRA embedding layer rank = 2

Figure 18: P-probing accuracies on the *minority group* with or without **celebrity** data. Each row represents a different pretrained model using its associated knowledge augmentation on the bioS data (*with or without celebrity data*). For every $i \in \{0, 1, \ldots, 5\}$ and $field \in \{\text{bmonth,bcity},\ldots\}$, the column labeled "$i$-$field$" shows the accuracy when predicting the first token / whole attribute of $field$ from the special position $i$, among individuals in the minority group.

| Model | QA mean acc | QA b_date | QA b_city | QA univ | QA major | QA c_name | QA c_city | Q-first b_month | Q-first b-city | Q-first univ | Q-first major | Q-first c_name | Q-first c_city | Q-whole b-city | Q-whole univ | Q-whole major | Q-whole c_name | Q-whole c_city |
|---|---|---|---|---|---|---|---|---|---|---|---|---|---|---|---|---|---|---|
| baseline | 2.7 | 0.0 | 0.5 | 0.3 | 1.0 | 0.4 | 13.7 | 8.3 | 2.5 | 37.0 | 4.0 | 1.5 | 14.8 | 0.5 | 0.3 | 1.0 | 0.4 | 13.7 |
| bioS single | 9.7 | 33.5 | 6.3 | 2.3 | 4.0 | 1.1 | 13.8 | 63.4 | 1.9 | 37.5 | 3.1 | 0.3 | 13.1 | 1.1 | 0.3 | 1.4 | 0.3 | 11.6 |
| bioS single + permute1 | 4.4 | 0.5 | 3.3 | 2.4 | 5.0 | 3.5 | 13.7 | 10.2 | 1.2 | 37.5 | 2.9 | 0.7 | 12.8 | 0.5 | 0.4 | 1.5 | 0.6 | 12.0 |
| bioS single + permute1 + CEL | 86.8 | 98.3 | 96.8 | 90.7 | 90.2 | 71.7 | 80.1 | 98.2 | 97.6 | 96.0 | 93.5 | 80.5 | 80.3 | 91.7 | 58.4 | 85.3 | 80.2 | 79.4 |
| bioR single | 10.0 | 25.1 | 13.9 | 2.4 | 5.5 | 2.0 | 14.1 | 28.0 | 1.4 | 35.7 | 5.4 | 0.6 | 10.5 | 1.1 | 0.7 | 2.7 | 0.4 | 9.5 |
| bioR single + wiki | 7.3 | 18.4 | 5.2 | 2.6 | 4.3 | 1.8 | 14.1 | 11.0 | 1.0 | 35.9 | 5.3 | 0.6 | 11.2 | 0.7 | 0.3 | 2.7 | 0.6 | 10.4 |
| bioR single + CEL | 76.3 | 94.3 | 85.3 | 82.9 | 79.4 | 67.0 | 56.6 | 97.9 | 89.8 | 89.4 | 84.2 | 75.2 | 57.9 | 87.0 | 71.5 | 80.3 | 73.5 | 59.1 |
| bioR single + fullname | 40.9 | 61.3 | 44.8 | 36.6 | 47.9 | 37.4 | 19.9 | 65.9 | 28.4 | 44.9 | 50.0 | 24.8 | 20.7 | 26.0 | 16.7 | 41.4 | 23.9 | 20.0 |
| bioR single + fullname + wiki | 27.7 | 39.3 | 33.3 | 25.8 | 33.5 | 22.2 | 14.6 | 61.1 | 22.5 | 40.4 | 32.8 | 13.5 | 14.7 | 21.4 | 12.0 | 25.8 | 11.8 | 13.7 |
| bioR single + fullname + CEL | 82.2 | 97.2 | 89.7 | 87.3 | 88.9 | 77.8 | 60.2 | 98.2 | 95.7 | 92.2 | 92.3 | 87.3 | 69.6 | 92.7 | 70.9 | 88.8 | 85.7 | 71.5 |

Figure 19: Q-probing accuracies on the *minority group* with or without **celebrity** data. Each row denotes a pretrained model with its specific knowledge augmentation. The left block reiterates QA finetune accuracies on the minority group (same as Figure 17). The middle showcases Q-probing accuracies on the first-token prediction for the six attributes of individuals in the minority group, and the right focuses on Q-probing for the "whole-attribute" prediction. Recall we have used a LoRA embedding rank 16 for the bioS data and rank 32 for the bioR data (see Appendix F).

# H  Details on BERT Experiment

Recall that GBERT is a bi-directional variant of GPT2, using the same tokenizer, as detailed in Appendix B. It is similar to BERT, but its architecture closely resembles GPT2 for a more direct comparison. We use GBERT for the following tasks: (1) BIO pretrain, (2) BIO+QA mixed training, (3) QA finetune from BIO pretrain, and (4) Q-probing from BIO pretrain. Since we only apply GBERT to the bioS data family to demonstrate a negative result, we utilize the same architecture size as GPT2-small.

For BIO pretrain and BIO+QA mixed training, we use the AdamW optimizer with weight decay 0.1, $\varepsilon = 10^{-6}$, an initial learning rate of 0.0003, a 1000-step linear warmup, and cosine learning rate decay (from 0.0003 to 0.00003). We use a batch size of 96 for 150000 training steps on the bioS dataset. This is *twice the training time* compared to the 80000 steps used for GPT2 small on the same dataset, as we are presenting a negative result on GBERT. For BIO+QA mixed training, we tested both $QA_r = 0.2$ and $QA_r = 0.8$ and report the best test accuracy.

For QA finetune, we tested four LoRA variants and report their best accuracy.[28] We use the AdamW optimizer with weight decay 0.01 and an initial learning rate of 0.0003 for all tasks, with linear learning rate decay (down to 0). We use a batch size of 48 for 50000 training steps.

For Q-probing, we use the AdamW optimizer with $\varepsilon = 10^{-6}$, weight decay 0.3, an initial learning rate of 0.001, no warmup, linear learning rate decay (down to 0), a batch size of 200, and 30000 training steps. This is identical to the procedure outlined in Appendix F.

All of our results were presented in the same Figure 9.

---

[28]Specifically, we tested rank-8 or rank-32 update on the query/value matrices, and rank-128 update or full finetuning on the embedding layer.